\pdfoutput=1
\documentclass[letterpaper, 10 pt, conference]{ieeeconf}  

\IEEEoverridecommandlockouts                              

\usepackage{siunitx}
\usepackage{graphics} 
\usepackage{epsfig} 
\usepackage{dblfloatfix}
\usepackage{tablefootnote}

\title{\LARGE \bf
Domain Randomization for Transferring Deep Neural Networks from Simulation to the Real World}
\author{Josh Tobin$^{1}$, Rachel Fong$^{2}$, Alex Ray$^{2}$, Jonas Schneider$^{2}$, Wojciech Zaremba$^{2}$, Pieter Abbeel$^{3}$
\thanks{$^{1}$OpenAI and UC Berkeley EECS,
        {\tt\small josh@openai.com}}%
\thanks{$^{2}$OpenAI,
        {\tt\small \{rfong, aray, jonas, woj\}@openai.com}}%
\thanks{$^{3}$OpenAI, UC Berkeley EECS \& ICSI,
        {\tt\small pieter@openai.com}}
}
\begin{document}

\maketitle
\thispagestyle{empty}
\pagestyle{empty}

\begin{abstract}

Bridging the `reality gap' that separates simulated robotics from experiments on hardware could accelerate robotic research through improved data availability. This paper explores \emph{domain randomization},
a simple technique for training models on simulated images that transfer to real images by randomizing rendering in the simulator. With enough variability in the simulator, the real world may appear to the model as just another variation. We focus on the task of object localization, which is a stepping stone to general robotic manipulation skills. We find that it is possible to train a real-world object detector that is accurate to \SI{1.5}{\centi\meter} and robust to distractors and partial occlusions using only data from a simulator with non-realistic random textures.
To demonstrate the capabilities of our detectors, we show they can be used to perform grasping in a cluttered environment. To our knowledge, this is the first successful transfer of a deep neural network trained \emph{only} on simulated RGB images (without pre-training on real images) to the real world for the purpose of robotic control.

\end{abstract}

\section{INTRODUCTION}

Performing robotic learning in a physics simulator could accelerate the impact of machine learning on robotics by allowing faster, more scalable, and lower-cost data collection than is possible with physical robots.  Learning in simulation is especially promising for building on recent results using deep reinforcement learning to achieve human-level performance on tasks like Atari \cite{mnih2015human} and robotic control \cite{levine2016end, schulman2015trust}.
Deep reinforcement learning employs random exploration, which can be dangerous on physical hardware. It often requires hundreds of thousands or millions of samples \cite{mnih2015human}, which could take thousands of hours to collect, making it impractical for many applications. Ideally, we could learn policies that encode complex behaviors entirely in simulation and successfully run those policies on physical robots with minimal additional training. 

Unfortunately, discrepancies between physics simulators and the real world make transferring behaviors from simulation challenging. {\it System identification}, the process of tuning the parameters of the simulation to match the behavior of the physical system, is time-consuming and error-prone. Even with strong system identification, the real world has {\it unmodeled physical effects} like nonrigidity, gear backlash, wear-and-tear, and fluid dynamics that are not captured by current physics simulators. Furthermore, {\it low-fidelity simulated sensors} like image renderers are often unable to reproduce the richness and noise produced by their real-world counterparts. These differences, known collectively as the \emph{reality gap}, form the barrier to using simulated data on real robots.

\begin{figure}[t]
    \centering
    \includegraphics[width=1.0\linewidth]{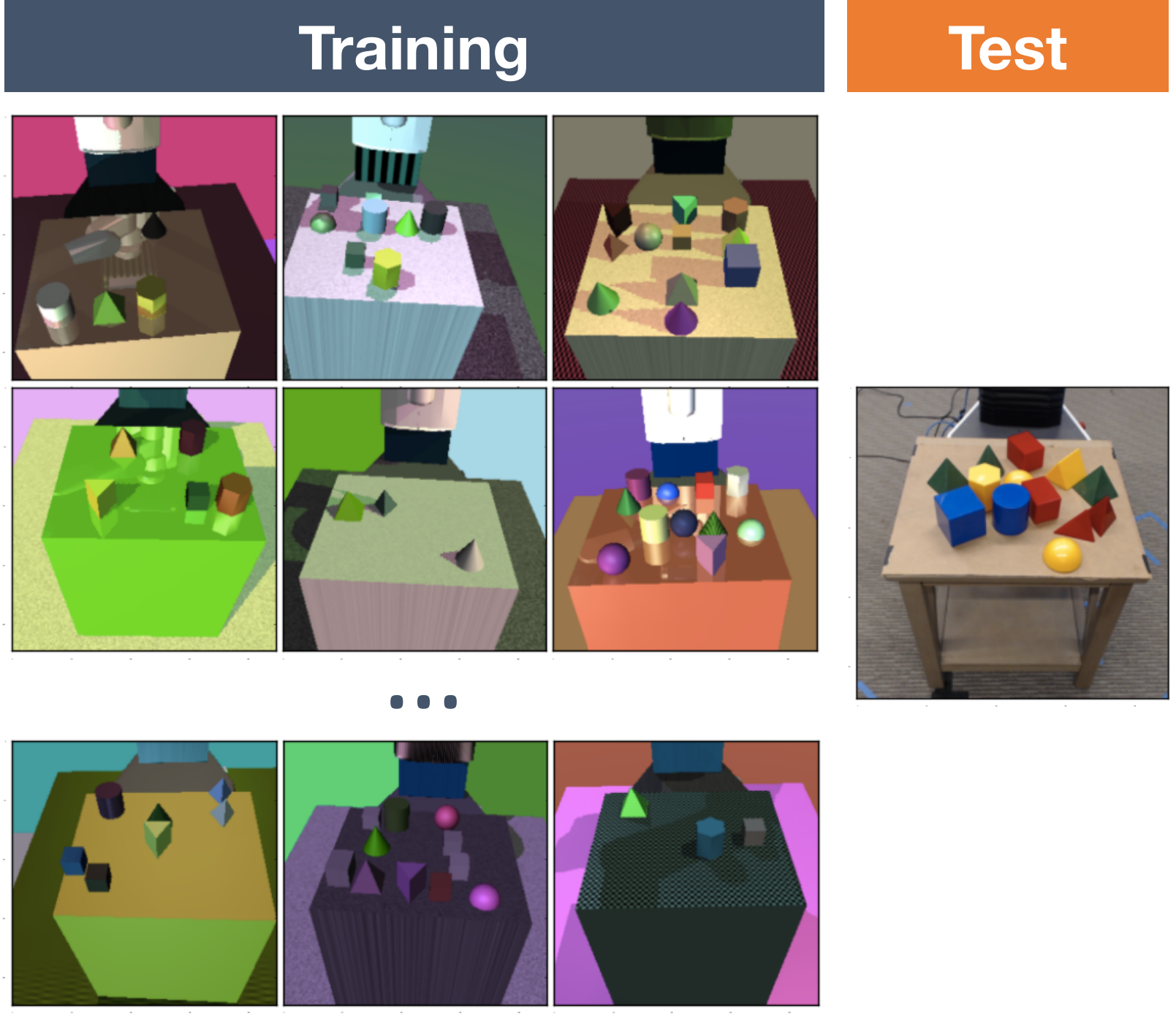}
    \caption{Illustration of our approach. An object detector is trained on hundreds of thousands of low-fidelity rendered images with random camera positions, 
    lighting conditions, object positions, and non-realistic textures. At test time, the same detector is used in the real world with no additional training.}
    \label{fig:main_fig}
\end{figure}

This paper explores {\it domain randomization}, a simple but promising method for addressing the reality gap. Instead of training a model on a single simulated environment, we randomize the simulator to expose the model to a wide range of environments at training time. The purpose of this work is to test the following hypothesis: if the variability in simulation is significant enough, models trained in simulation will generalize to the real world with no additional training. 

Though in principle domain randomization could be applied to any component of the reality gap, we focus on the challenge of transferring from low-fidelity simulated camera images.
Robotic control from camera pixels is attractive due to the low cost of cameras and the rich data they provide, but challenging because it involves processing high-dimensional input data. Recent work has shown that supervised learning with deep neural networks is a powerful tool for learning generalizable representations from high-dimensional inputs \cite{lecun2015deep}, but deep learning relies on a large amount of labeled data.
Labeled data is difficult to obtain in the real world for precise robotic manipulation behaviors, but it is easy to generate in a physics simulator.

We focus on the task of training a neural network to detect the location of an object. Object localization from pixels is a well-studied problem in robotics, and state-of-the-art methods employ complex, hand-engineered image processing pipelines (e.g., \cite{collet2010efficient}, \cite{collet2011moped}, \cite{tang2012textured}). This work is a first step toward the goal of using deep learning to improve the accuracy of object detection pipelines. Moreover, we see sim-to-real transfer for object localization as a stepping stone to transferring general-purpose manipulation behaviors.

We find that for a range of geometric objects, we are able to train a detector that is accurate to around \SI{1.5}{\centi\meter} in the real world using only simulated data rendered with simple, algorithmically generated textures. Although previous work demonstrated the ability to perform robotic control using a neural network pretrained on ImageNet 
and fine-tuned on randomized rendered pixels \cite{sadeghi2016cad}, this paper provides the first demonstration that domain randomization can be useful for robotic tasks requiring precision. We also provide an ablation study of the impact of different choices of randomization and training method on the success of transfer. We find that with a sufficient number of textures, pre-training the object detector using real images is unnecessary. To our knowledge, this is the first successful transfer of a deep neural network trained \emph{only} on simulated RGB images to the real world for the purpose of robotic control.

\section{RELATED WORK}

\subsection{Object detection and pose estimation for robotics}
Object detection and pose estimation for robotics is a well-studied problem in the literature (see, e.g., \cite{collet2009object}, \cite{collet2011moped}, \cite{collet2010efficient},  \cite{ekvall2005object}, \cite{tang2012textured}, \cite{wunsch1997real}, \cite{zickler2006detection}). Recent approaches typically involve offline construction or learning of a 3D model of objects in the scene (e.g., a full 3D mesh model \cite{tang2012textured} or a 3D metric feature representation \cite{collet2011moped}). At test time, features from the test data (e.g., Scale-Invariant Feature Transform [SIFT] features \cite{gordon2006and} or color co-occurrence histograms \cite{ekvall2005object}) are matched with the 3D models (or features from the 3D models). For example, a black-box nonlinear optimization algorithm can be used to minimize the re-projection error of the SIFT points from the object model and the 2D points in the test image \cite{collet2009object}. Most successful approaches rely on using multiple camera frames \cite{collet2010efficient} or depth information \cite{tang2012textured}. There has also been some success with only monocular camera images  \cite{collet2009object}. 

Compared to our method, traditional approaches require less extensive training and take advantage of richer sensory data, allowing them to detect the full 3D pose of objects (position and orientation) without any assumptions about the location or size of the surface on which the objects are placed. However, our approach avoids the challenging problem of 3D reconstruction, and employs a simple, easy to implement deep learning-based pipeline that may scale better to more challenging problems.

\subsection{Domain adaptation}
The computer vision community has devoted significant study to the problem of adapting vision-based models trained in a source domain to a previously unseen target domain (see, e.g., \cite{duan2012learning}, \cite{Hoffman_NIPS2014}, \cite{hoffman2013efficient}, \cite{kulis2011you}, \cite{long2015learning}, \cite{mansour2009domain}, \cite{yang2007cross}). A variety of approaches have been proposed, including re-training the model in the target domain (e.g., \cite{yosinski2014transferable}), adapting the weights of the model based on the statistics of the source and target domains (e.g., \cite{li2016revisiting}), learning invariant features between domains (e.g., \cite{tzeng2014deep}), and learning a mapping from the target domain to the source domain (e.g., \cite{taigman2016unsupervised}). Researchers in the reinforcement learning community have also studied the problem of domain adaptation by learning invariant feature representations \cite{gupta2016feature}, adapting pretrained networks \cite{rusu2016progressive}, and other methods. See \cite{gupta2016feature} for a more complete treatment of domain adaptation in the reinforcement learning literature.

In this paper we study the possibility of transfer from simulation to the real world \emph{without} performing domain adaptation. 

\subsection{Bridging the reality gap}
Previous work on leveraging simulated data for physical robotic experiments explored several strategies for bridging the reality gap. 

One approach is to make the simulator closely match the physical reality by performing system identification and using high-quality rendering. Though using realistic RGB rendering alone has had limited success for transferring to real robotic tasks \cite{james20163d}, incorporating realistic simulation of depth information can allow models trained on rendered images to transfer reasonably well to the real world \cite{planche2017depthsynth}. Combining data from high-quality simulators with other approaches like fine-tuning can also reduce the number of labeled samples required in the real world \cite{richter2016playing}. 

Unlike these approaches, ours allows the use of low-quality renderers optimized for speed and not carefully matched to real-world textures, lighting, and scene configurations. 

Other work explores using domain adaptation techniques to bridge the reality gap. It is often faster to fine-tune a controller learned in simulation than to learn from scratch in the real world \cite{cutler2015efficient, kolter2007learning}. In \cite{ghadirzadeh2017deep}, the authors use a variational autoencoder trained on simulated data to encode trajectories of motor outputs corresponding to a desired behavior type (e.g., reaching, grasping) as a low-dimensional latent code. A policy is learned on real data mapping features to distributions over latent codes. The learned policy overcomes the reality gap by choosing latent codes that correspond to the desired physical behavior via exploration. 

Domain adaptation has also been applied to robotic vision. Rusu et al. \cite{rusu2016sim} explore using the progressive network architecture to adapt a model that is pre-trained on simulated pixels, and find it has better sample efficiency than fine-tuning or training in the real-world alone. In \cite{tzeng2016adapting}, the authors explore learning a correspondence between domains that allows the real images to be mapped into a space understood by the model. While both of the preceding approaches require reward functions or labeled data, which can be difficult to obtain in the real world, Mitash and collaborators \cite{mitash2017self} explore pretraining an object detector using realistic rendered images with randomized lighting from 3D models to bootstrap an automated learning learning process that does not require manually labeling data and uses only around 500 real-world samples. 

A related idea, {\it iterative learning control}, employs real-world data to improve the dynamics model used to determine the optimal control behavior, rather than using real-world data to improve the controller directly. Iterative learning control starts with a dynamics model, applies the corresponding control behavior on the real system, and then closes the loop by using the resulting data to improve the dynamics model. Iterative learning control has been applied to a variety of robotic control problems, from model car control
(e.g., \cite{abbeel2006using} and \cite{cutler2014reinforcement}) to surgical robotics (e.g., \cite{van2010superhuman}). 

Domain adaptation and iterative learning control are important tools for addressing the reality gap, but in contrast to these approaches, ours requires no additional training on real-world data. Our method can also be combined easily with most domain adaptation techniques. 

Several authors have previously explored the idea of using domain randomization to bridge the reality gap. 

In the context of physics adaptation, Mordatch and collaborators \cite{mordatch2015ensemble} show that training a policy on an ensemble of dynamics models can make the controller robust to modeling error and improve transfer to a real robot. Similarly, in \cite{antonova2017reinforcement}, the authors train a policy to pivot a tool held in the robot's gripper in a simulator with randomized friction and action delays, and find that it works in the real world and is robust to errors in estimation of the system parameters. 

Rather than relying on controller robustness, Yu et al. \cite{yu2017preparing} use a model trained on varied physics to perform system identification using online trajectory data, but their approach is not shown to succeed in the real world. Rajeswaran et al. \cite{rajeswaran2016epopt} explore different training strategies for learning from an ensemble of models, including adversarial training and adapting the ensemble distribution using data from the target domain, but also do not demonstrate successful real-world transfer. 

Researchers in computer vision have used 3D models as a tool to improve performance on real images since the earliest days of the field (e.g., \cite{nevatia1977description}, \cite{lowe1987three}). More recently, 3D models have been used to augment training data to aid transferring deep neural networks between datasets and prevent over-fitting on small datasets for tasks like viewpoint estimation \cite{su2015render} and object detection \cite{sun2014virtual}, \cite{movshovitz2016useful}. Recent work has explored using only synthetic data for training 2D object detectors (i.e., predicting a bounding box for objects in the scene). In \cite{peng2015learning}, the authors find that by pretraining a network on ImageNet and fine-tuning on synthetic data created from 3D models, better detection performance on the PASCAL dataset can be achieved than training with only a few labeled examples from the real dataset. 

In contrast to our work, most object detection results in computer vision use realistic textures, but do not create coherent 3D scenes. Instead, objects are rendered against a solid background or a randomly chosen photograph. As a result, our approach allows our models to understand the 3D spatial information necessary for rich interactions with the physical world.

Sadeghi and Levine's work \cite{sadeghi2016cad} is the most similar to our own. The authors demonstrate that a policy mapping images to controls learned in a simulator with varied 3D scenes and textures can be applied successfully to real-world quadrotor flight. However, their experiments -- collision avoidance in hallways and open spaces -- do not demonstrate the ability to deal with high-precision tasks. Our approach also does not rely on precise camera information or calibration, instead randomizing the position, orientation, and field of view of the camera in the simulator. Whereas their approach chooses textures from a dataset of around $200$ pre-generated materials, most of which are realistic, our approach is the first to use only non-realistic textures created by a simple random generation process, which allows us to train on hundreds of thousands (or more) of unique texturizations of the scene. 

\section{METHOD}
Given some objects of interest $\{s_i\}_i$, our goal is to train an object detector $d(I_0)$ that maps a single monocular camera frame $I_0$ to the Cartesian coordinates $\{(x_i, y_i, z_i)\}_i$ of each object. In addition to the objects of interest, our scenes sometimes contain distractor objects that must be ignored by the network. Our approach is to train a deep neural network in simulation using domain randomization. The remainder of this section describes the specific domain randomization and neural network training methodology we use.

\subsection{Domain randomization}
The purpose of domain randomization is to provide enough simulated variability at training time such that at test time the model is able to generalize to real-world data. We randomize the following aspects of the domain for each sample used during training:
\begin{itemize}
\item Number and shape of distractor objects on the table 
\item Position and texture of all objects on the table
\item Textures of the table, floor, skybox, and robot
\item Position, orientation, and field of view of the camera
\item Number of lights in the scene
\item Position, orientation, and specular characteristics of the lights
\item Type and amount of random noise added to images 
\end{itemize}

Since we use a single monocular camera image from an uncalibrated camera to estimate object positions, we fix the height of the table in simulation, effectively creating a 2D pose estimation task. Random textures are chosen among the following:
\begin{enumerate}
\item[(a)] A random RGB value
\item[(b)] A gradient between two random RGB values
\item[(c)] A checker pattern between two random RGB values
\end{enumerate}
The textures of all objects are chosen uniformly at random -- the detector does not have access to the color of the object(s) of interest at training time, only their size and shape. We render images using the MuJoCo Physics Engine's \cite{todorov2012mujoco} built-in renderer. This renderer is not intended to be photo-realistic, and physically plausible choices of textures and lighting are not needed. 

Between $0$ and $10$ distractor objects are added to the table in each scene. Distractor objects on the floor or in the background are unnecessary, despite some clutter (e.g., cables) on the floor in our real images.

Our method avoids calibration and precise placement of the camera in the real world by randomizing characteristics of the cameras used to render images in training. We manually place a camera in the simulated scene that approximately matches the viewpoint and field of view of the real camera. Each training sample places the camera randomly within a $(10 \times 5 \times 10)$ cm box around this initial point. The viewing angle of the camera is calculated analytically to point at a fixed point on the table, and then offset by up to $0.1$ radians in each direction. The field of view is also scaled by up to \SI{5}{\percent} from the starting point.

\subsection{Model architecture and training}
\begin{figure}[h!]
    \centering
    \includegraphics[width=1.0\linewidth]{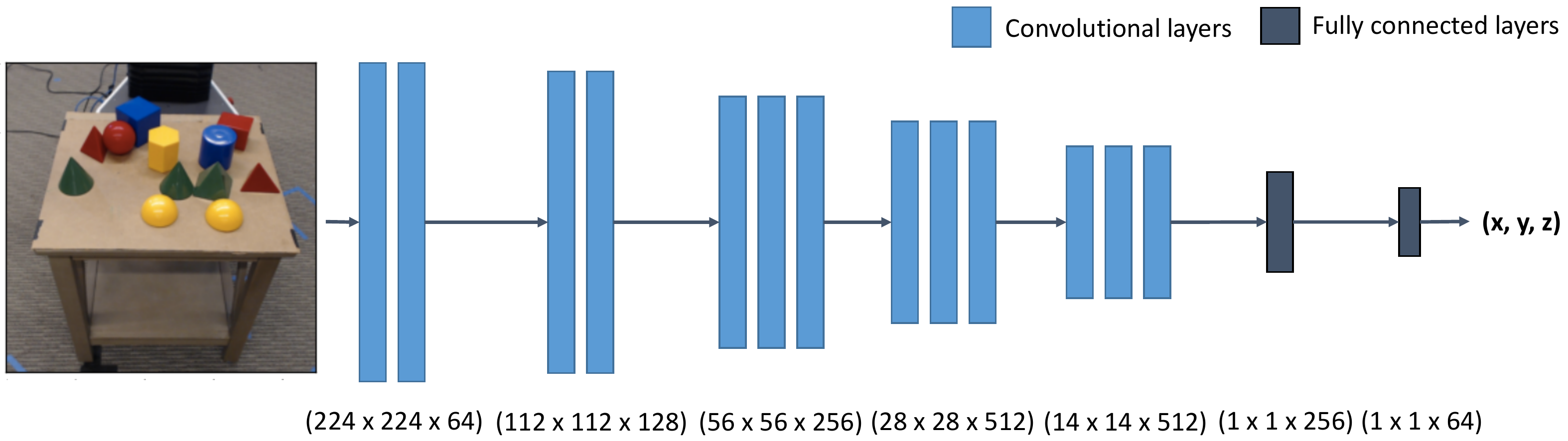}
    \caption{The model architecture used in our experiments. Each vertical bar corresponds to a layer of the model. ReLU nonlinearities are used throughout, and max pooling occurs between each of the groupings of convolutional layers. The input is an image from an external webcam downsized to $(224 \times 224)$ and the output of the network predicts the $(x, y, z)$ coordinates of object(s) of interest.}
    \label{fig:model_architecture}
\end{figure}
We parametrize our object detector with a deep convolutional neural network. In particular, we use a modified version the VGG-16 architecture \cite{simonyan2014very} shown in Figure \ref{fig:model_architecture}. We chose this architecture because it performs well on a variety of computer vision tasks, and because it has a wide availability of pretrained weights. We use the standard VGG convolutional layers, but use smaller fully connected layers of sizes $256$ and $64$ and do not use dropout. For the majority of our experiments, we use weights obtained by pretraining on ImageNet to initialize the convolutional layers, which we hypothesized would be essential to achieving transfer. In practice, we found that using random weight initialization works as well in most cases.

We train the detector through stochastic gradient descent on the $L_2$ loss between the object positions estimated by the network and the true object positions using the Adam optimizer \cite{kingma2014adam}. We found that using a learning rate of around $1\mathrm{e}{-4}$ (as opposed to the standard $1\mathrm{e}{-3}$ for Adam) improved convergence and helped avoid a common local optimum, mapping all objects to the center of the table.

\section{EXPERIMENTS}

\subsection{Experimental Setup}
\begin{figure}[h!]
    \centering
    \includegraphics[width=0.8\linewidth]{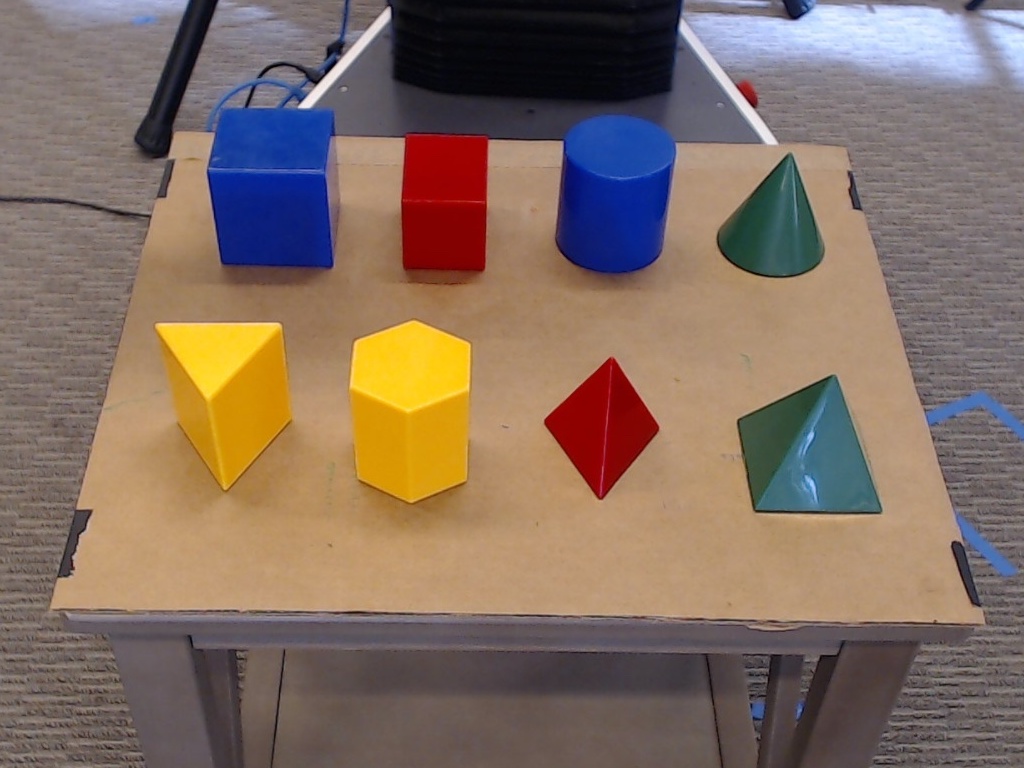}
    \caption{The geometric objects used in our experiments.}
    \label{}
\end{figure}
We evaluated our approach by training object detectors for each of eight geometric objects. We constructed mesh representations for each object to render in the simulator. Each training sample consists of (a) a rendered image of the object and one or more distractors (also from among the geometric object set) on a simulated tabletop and (b) a label corresponding to the Cartesian coordinates of the center of mass of the object in the world frame. 

For each experiment, we performed a small hyperparameter search, evaluating combinations of two learning rates ($1 \mathrm{e}{-4}$ and $2\mathrm{e}{-4}$) and three batch sizes ($25$, $50$, and $100$). We report the performance of the best network.

The goals of our experiments are:
\begin{enumerate}
\item[(a)] Evaluate the localization accuracy of our trained detectors in the real world, including in the presence of distractor objects and partial occlusions
\item[(b)] Assess which elements of our approach are most critical for achieving transfer from simulation to the real world
\item [(c)] Determine whether the learned detectors are accurate enough to perform robotic manipulation tasks
\end{enumerate}

\begin{table}[ht!]
\label{table:maintable}
\caption{}
\scriptsize
\centering
\renewcommand{\arraystretch}{1.3}
\begin{tabular}{|c|c|c|c|}
\hline
\multicolumn{4}{|c|}{\bf Detection error for various objects, cm} \\ \hline 
 Evaluation type & Object only & Distractors & Occlusions  \\ \hline
Cone & $1.3 \pm 1.1$\footnotemark & $1.5 \pm 1.0$  & $1.4 \pm 0.6$ \\ \hline
Cube & $1.3 \pm 0.6$  & $1.8 \pm 1.2$  & $1.4 \pm 0.6^{\text{1}}$ \\ \hline
Cylinder & $1.1 \pm 0.9^{\text{1}}$  & $1.9 \pm 2.8$ & $1.9 \pm 2.9$ \\ \hline
Hexagonal Prism & $0.7 \pm 0.5$  & $0.6 \pm 0.3^{\text{1}}$  &$ 1.0 \pm 1.0^{\text{1}}$  \\ \hline
Pyramid & $0.9 \pm 0.3^{\text{1}}$  & $1.0 \pm 0.5^{\text{1}}$  &$1.1 \pm 0.7^{\text{1}}$  \\ \hline
Rectangular Prism & $1.3 \pm 0.7$ & $1.2 \pm 0.4^{\text{1}}$  & $0.9 \pm 0.6$  \\ \hline
Tetrahedron & $0.8 \pm 0.4^{\text{1}}$  & $1.0 \pm 0.4^{\text{1}}$ & $3.2 \pm 5.8$ \\ \hline
Triangular Prism & $0.9 \pm 0.4^{\text{1}}$ & $0.9 \pm 0.4^{\text{1}}$ & $1.9 \pm 2.2$ \\ \hline
\end{tabular}
\end{table}
\footnotetext{Categories for which the best final performance was achieved for detector trained from scratch.}
\subsection{Localization accuracy}

To evaluate the accuracy of learned detectors in the real world, we captured $480$ webcam images of one or more geometric objects on a table at a distance of \SIrange{70}{105}{\centi\meter} from the camera. The camera position remains constant across all images. We did not control for lighting conditions or the rest of the scene around the table (e.g., all images contain part of the robot and tape and wires on the floor). We measured ground truth positions for a single object per image by aligning the object on a grid on the tabletop. Each of the eight geometric objects has 60 labeled images in the dataset: $20$ with the object alone on the table, $20$ in which one or more distractor objects are present on the table, and $20$ in which the object is also partially occluded by another object. 

Table I summarizes the performance of our models on the test set. Our object detectors are able to localize objects to within \SI{1.5}{\centi\meter} (on average) in the real world and perform well in the presence of clutter and partial occlusions. Though the accuracy of our trained detectors is promising, note that they are still over-fitting\footnote{Overfitting in this setting is more subtle than in the standard supervised learning where train and test data come from the same distribution.  In the standard supervised learning setting overfitting can be avoided by using a hold-out set during training.  We do apply this idea to ensure that we are not overfitting on the simulated data.  However, since our goal is to learn from training data originated in the simulator and generalize to test data originated from the real world, we assume to not have any real world data available during training. Therefore no validation on real data can be done during training.}  the simulated training data, where error is \SIrange{0.3}{0.5}{\centi\meter}. Even with over-fitting, the accuracy is comparable at a similar distance to the translation error in traditional techniques for pose estimation in clutter from a single monocular camera frame \cite{collet2011moped} that use higher-resolution images.

\subsection{Ablation study}
To evaluate the importance of different factors of our training methodology, we assessed the sensitivity of the algorithm to the following:
\begin{itemize}
\item Number of training images
\item Number of unique textures seen in training
\item Use of random noise in pre-processing
\item Presence of distractors in training
\item Randomization of camera position in training
\item Use of pre-trained weights in the detection model
\end{itemize}

We found that the method is at least somewhat sensitive to all of the factors except the use of random noise. 

\begin{figure}[h!]
    \centering
    \includegraphics[width=1.0\linewidth]{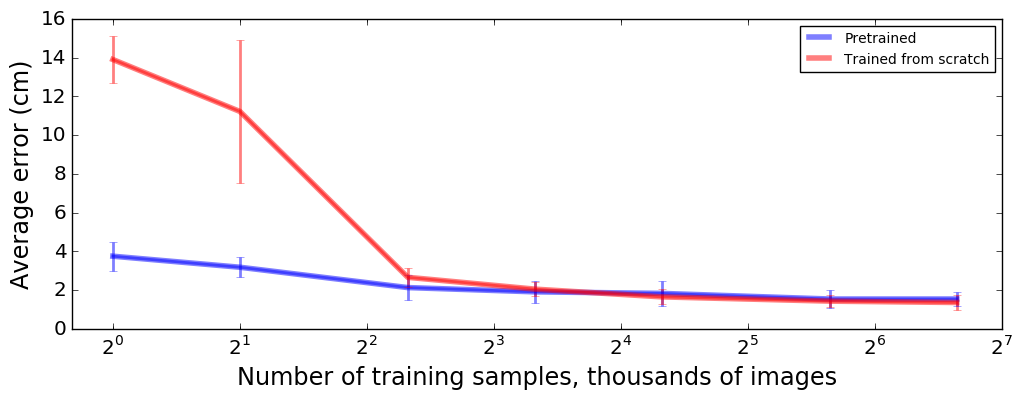}
    \caption{Sensitivity of test error on real images to the number of simulated training examples used. Each training example corresponds to a single labeled example of an object on the table with between 0 and 10 distractor objects. Lighting and all textures are randomized between iterations. }
    \label{fig:amt_data}
\end{figure}
Figure \ref{fig:amt_data} shows the sensitivity to the number of training samples used for pre-trained models and models trained from scratch. Using a pre-trained model, we are able to achieve relatively accurate real-world detection performance with as few as $5,000$ training samples, but performance improves up to around $50,000$ samples.

Figure \ref{fig:amt_data} also compares to the performance of a model trained from scratch (i.e., without using pre-trained ImageNet weights). Our hypothesis that pre-training would be essential to generalizing to the real world proved to be false. With a large amount of training data, random weight initialization can achieve nearly the same performance in transferring to the real world as does pre-trained weight initialization. The best detectors for a given object were often those initialized with random weights. However, using a pre-trained model can significantly improve performance when less training data is used. 

\begin{figure}[h!]
    \centering
    \includegraphics[width=1.0\linewidth]{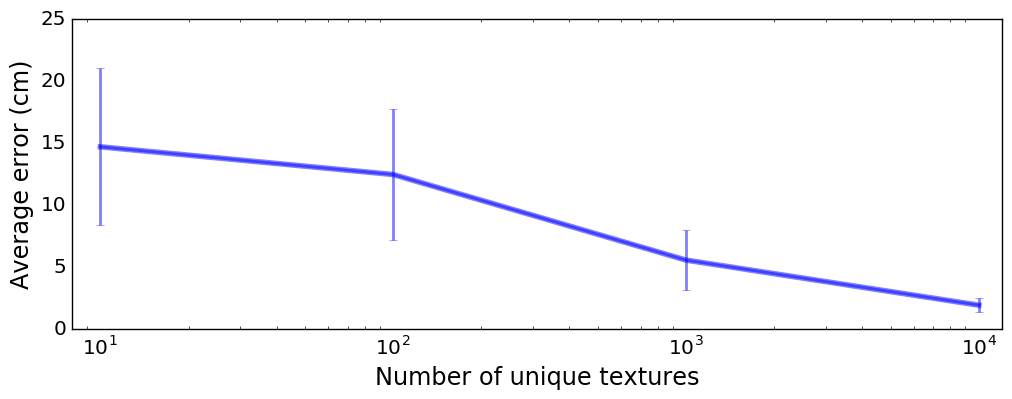}
    \caption{Sensitivity to amount of texture randomization. In each case, the detector was trained using $10,000$ random object positions and combinations of distractors, but only the given number of unique texturizations and lighting conditions were used.}
    \label{fig:amt_textures}
\end{figure}

Figure \ref{fig:amt_textures} shows the sensitivity to the number of unique texturizations of the scene when trained on a fixed number ($10,000$) of training examples. We found that performance degrades significantly when fewer than $1,000$ textures are used, indicating that for our experiments, using a large number of random textures (in addition to random distractors and object positions) is necessary to achieving transfer. Note that when $1,000$ random textures are used in training, the performance using $10,000$ images is comparable to that of using only $1,000$ images, indicating that in the low data regime, texture randomization is more important than randomization of object positions.\footnote{Note the total number of textures is higher than the number of training examples in some of these experiments because every scene has many surfaces, each with its own texture.}

Table II examines the performance of the algorithm when random noise, distractors, and camera randomization are removed in training. Incorporating distractors during training appears to be critical to resilience to distractors in the real world. Randomizing the position of the camera also consistently provides a slight accuracy boost, but reasonably high accuracy is achievable without it. Adding noise during pretraining appears to have a negligible effect. In practice, we found that adding a small amount of random noise to images at training time improves convergence and makes training less susceptible to local minima.

\begin{table}[h]
\label{table:ablation}
\caption{}
\scriptsize
\centering
\renewcommand{\arraystretch}{1.3}
\begin{tabular}{|c|c|c|c|}
\hline
\multicolumn{4}{|c|}{\bf Average detection error on geometric shapes by method, cm\footnotemark} \\ \hline 
 Evaluation &  \multicolumn{3}{|c|}{\it Real images} \\ 
\cline{2-4}
type & Object only & Distractors & Occlusions \\ \hline
Full method & $\bf{1.3 \pm 0.6}$  & $\bf{1.8 \pm 1.7}$  & $\bf{2.4 \pm 3.0}$   \\ \hline
No noise added & $1.4 \pm 0.7$  & $1.9 \pm 2.0$  & $\bf{2.4 \pm 2.8}$ \\ \hline
No camera randomization & $2.0 \pm 2.1$  & $2.4 \pm 2.3$  &$ 2.9 \pm 3.5$  \\ \hline
No distractors in training & $1.5 \pm 0.6$  & $7.2 \pm 4.5$  &$ 7.4 \pm 5.3$ \\ \hline

\end{tabular}
\end{table}

\subsection{Robotics experiments}
To demonstrate the potential of this technique for transferring robotic behaviors learned in simulation to the real world, we evaluated the use of our object detection networks for localizing an object in clutter and performing a prescribed grasp. For two of our most consistently accurate detectors, we evaluated the ability to pick up the detected object in 20 increasingly cluttered scenes using the positions estimated by the detector and off-the-shelf motion planning software \cite{moveit}. To test the robustness of our method to discrepancies in object distributions between training and test time, some of our test images contain distractors placed at orientations not seen during training (e.g., a hexagonal prism placed on its side). 

We deployed the pipeline on a Fetch robot \cite{wise2016fetch}, and found it was able to successfully detect and pick up the target object in 38 out of 40 trials, including in highly cluttered scenes with significant occlusion of the target object. Note that the trained detectors have no prior information about the color of the target object, only its shape and size, and are able to detect objects placed closely to other objects of the same color. 

To test the performance of our object detectors on real-world objects with non-uniform textures, we trained an object detector to localize a can of Spam from the YCB Dataset \cite{calli2015ycb}. At training time, the can was present on the table along with geometric object distractors. At test time, instead of using geometric object distractors, we placed other food items from the YCB set on the table. The detector was able to ignore the previously unseen distractors and pick up the target in 9 of 10 trials.

\footnotetext{Each of the models compared was trained with $20,000$ training examples}

\begin{figure}[t!]
    \centering
    \includegraphics[width=1.0\linewidth]{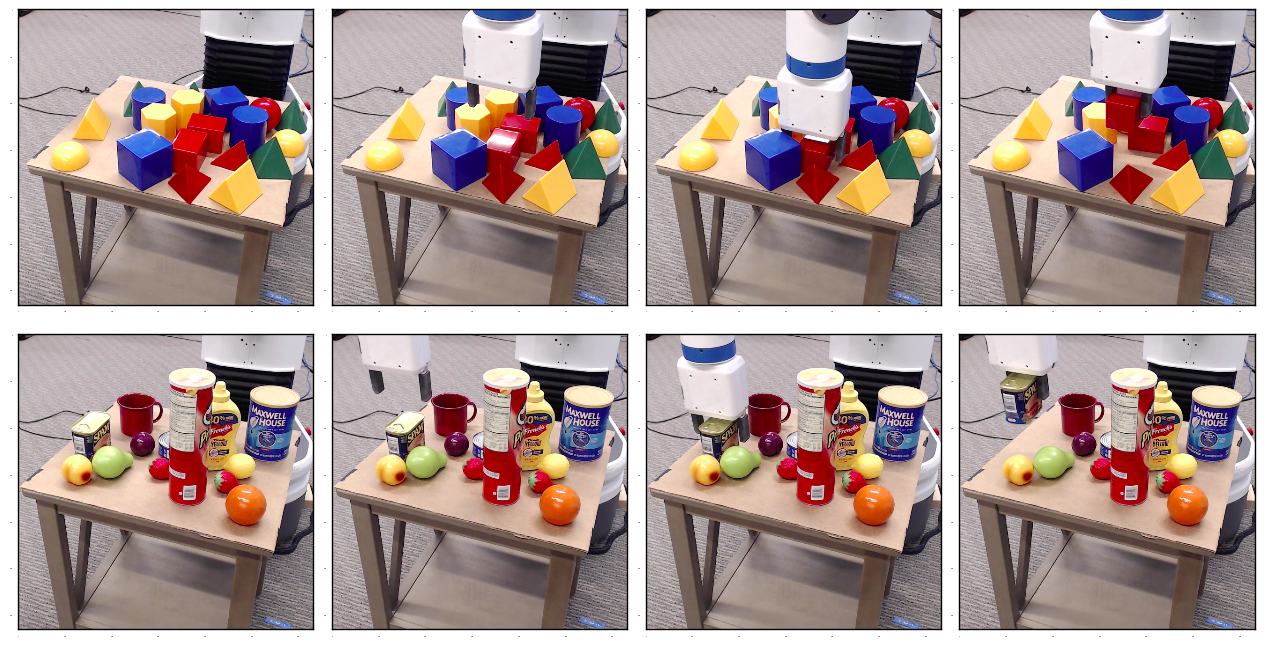}
    \caption{Two representative executions of grasping objects using vision learned in simulation only. The object detector network estimates the positions of the object of interest, and then a motion planner plans a simple sequence of motions to grasp the object at that location.}
    \label{fig:grasping}
\end{figure}

Figure \ref{fig:grasping} shows examples of the robot grasping trials. For videos, please visit the web page associated with this paper.\footnote{\url{https://sites.google.com/view/domainrandomization/}}

\section{CONCLUSION}
We demonstrated that an object detector trained only in simulation can achieve high enough accuracy in the real world to perform grasping in clutter. Future work will explore how to make this technique reliable and effective enough to perform tasks that require contact-rich manipulation or higher precision.

Future directions that could improve the accuracy of object detectors trained using domain randomization include:
\begin{itemize}
\item Using higher resolution camera frames
\item Optimizing model architecture choice
\item Introducing additional forms of texture, lighting, and rendering randomization to the simulation and training on more data
\item Incorporating multiple camera viewpoints, stereo vision, or depth information
\item Combining domain randomization with domain adaptation
\end{itemize}

Domain randomization is a promising research direction toward bridging the reality gap for robotic behaviors learned in simulation. Deep reinforcement learning may allow more complex policies to be learned in simulation through large-scale exploration and optimization, and domain randomization could be an important tool for making such policies useful on real robots.

\bibliographystyle{plain}

\section*{APPENDIX}

\subsection{Randomly generated samples from our method}
\begin{figure}[h]
    \centering
    \includegraphics[width=0.92\linewidth]{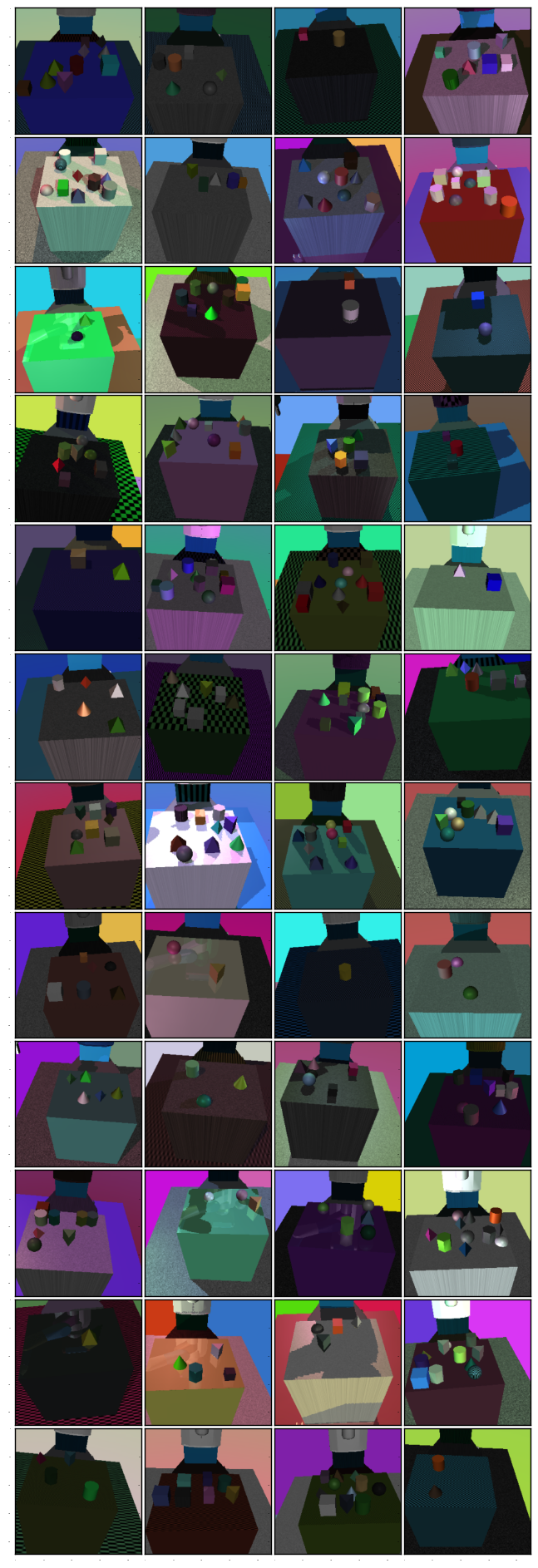}
    \caption{A selection of randomly textured scenes used in the training phase of our method}
    \label{fig:random_samples}
\end{figure}

Figure \ref{fig:random_samples} displays a selection of the images used during training for the object detectors detailed in the paper. 


\end{document}